%% file: main.tex
\documentclass{article} %

\usepackage[preprint]{neurips_2025}

\usepackage[utf8]{inputenc} %
\usepackage[T1]{fontenc}    %
\usepackage{xurl}            %
\usepackage{booktabs}       %
\usepackage{amsfonts}       %
\usepackage{nicefrac}       %
\usepackage{microtype}      %
\usepackage[dvipsnames, table]{xcolor}         %
\usepackage[position=bottom]{caption}
\usepackage{enumitem}
\usepackage[colorlinks=true,
            linkcolor=Blue,
            urlcolor=Blue,
            citecolor=Blue,
            anchorcolor=Blue]{hyperref}   
\usepackage{graphicx}
\usepackage{booktabs}
\usepackage{amsmath}
\usepackage{multirow}
\usepackage{makecell}
\usepackage{fontawesome5}
\usepackage{tcolorbox}  %
\usepackage{pifont}
\usepackage{amssymb}
\usepackage[textsize=tiny]{todonotes}

\definecolor{darkblue}{rgb}{0, 0, 0.5}
\hypersetup{colorlinks=true, citecolor=darkblue, linkcolor=darkblue, urlcolor=darkblue}

\definecolor{greenbox}{rgb}{0.83, 0.95, 0.85}
\definecolor{orangebox}{rgb}{0.97, 0.9, 0.7}
\definecolor{redbox}{rgb}{0.97, 0.83, 0.83}

\newtcbox{\greenbox}{on line, box align=base, colback=greenbox, colframe=white, size=fbox, arc=3pt, before upper=\strut, top=-2pt, bottom=-4pt, left=-2pt, right=-2pt, boxrule=0pt}
\newtcbox{\orangebox}{on line, box align=base, colback=orangebox, colframe=white, size=fbox, arc=3pt, before upper=\strut, top=-2pt, bottom=-4pt, left=-2pt, right=-2pt, boxrule=0pt}
\newtcbox{\redbox}{on line, box align=base, colback=redbox, colframe=white, size=fbox, arc=3pt, before upper=\strut, top=-2pt, bottom=-4pt, left=-2pt, right=-2pt, boxrule=0pt}

\newcommand{\gbox}[1]{{\greenbox{#1}}}

\newcommand{\rbox}[1]{{\redbox{#1}}}

\title{Cache Me If You Can: How Many KVs Do You Need for Effective Long-Context LMs?}

\author{Adithya Bhaskar$^*$, Alexander Wettig$^*$, Tianyu Gao, Yihe Dong, Danqi Chen \\
Princeton Language and Intelligence, Princeton University \\
\texttt{adithyab@princeton.edu, awettig@cs.princeton.edu}
}

\newcommand{\ours}{{PruLong}}

\begin{document}

\maketitle

\begin{abstract}
\input{sections/00_abstract}

\renewcommand{\thefootnote}{\fnsymbol{footnote}}
\footnotetext[1]{The first two authors contributed equally.}
\end{abstract}

\input{sections/01_intro}

\input{sections/02_background}

\input{sections/03_method}

\input{sections/04_experiments}

\input{sections/05_conclusion}

\bibliographystyle{plainnat}
\bibliography{ref}

\appendix

\newpage
\input{appendices/kv_metrics}
\input{appendices/objective}
\input{appendices/experiments}
\input{appendices/real_metrics}

\end{document}

%% file: sections/00_abstract.tex
\label{sec:abstract}

Language models handle increasingly long contexts for tasks such as book summarization, but this leads to growing memory costs for the key-value (KV) cache. 
Many prior works have proposed ways of discarding KVs from memory,
but their approaches are tailored to favorable settings,
obscuring caveats like high peak memory and performance degradation,
and a fair comparison between methods is difficult.
In this paper, we propose the \emph{KV footprint} as a unified metric, which accounts for both the amount of KV entries stored and their lifespan in memory. 
We evaluate methods based on the smallest footprint they attain while preserving performance in both long-context understanding and generation, with context lengths of up to 128K tokens.
This metric reveals the high peak memory of prior KV eviction methods.
One class of methods---\emph{post-fill eviction}---has a high footprint due to being incompatible with eviction during pre-filling. 
We adapt these methods to be able to evict KVs during pre-filling, achieving substantially lower KV footprints.
We then turn to \emph{recency eviction} methods, wherein we propose \ours{}, an end-to-end optimization method for learning which attention heads need to retain the full KV cache and which do not.
\ours{} saves memory while preserving long-context performance, achieving 12\% smaller KV footprint than prior methods while retaining performance in challenging recall tasks.
Our paper clarifies the complex tangle of long-context inference methods and paves the way for future development to minimize the KV footprint.\footnote{We  release our code publicly at \url{https://github.com/princeton-pli/PruLong}.}

%% file: sections/01_intro.tex
\section{Introduction}
\label{sec:intro}

The long-context abilities of language models (LMs) enable applications from summarizing books to reviewing thousands of lines of code, but serving these workloads remains challenging due to the extreme memory requirements of long-context inference.
In addition, recent developments like long chains of thought~\citep{deepseekai2025deepseekr1} point to emerging workloads with long output sequences---often in the tens of thousands of tokens. 
Since most LMs use the Transformer architecture~\citep{vaswani2017attention}, autoregressive decoding requires storing all previous attention states in the \emph{Key-Value (KV) cache}.
The size of the KV cache grows linearly with the input length, and a 128K-token prompt ($\approx$ the Llama 3 technical report) for Llama-3-70B~\citep{dubey2024llama} would allocate 42GB of KV memory---a significant resource requirement for long-context inference.

Many methods have been proposed to reduce the long-context memory footprint by removing ``unimportant'' entries from the KV cache, %
but comparing them in a fair setting is difficult.
The challenge arises from several factors:
\begin{enumerate}[topsep=0pt,parsep=0pt,partopsep=0pt, leftmargin=2em]
\item Methods like MInference~\citep{jiang2024minference} focus on efficiency in pre-filling (encoding the prompt), while others such as PyramidKV~\citep{cai2024pyramidkv} and SnapKV~\citep{li2024snapkv} focus on decoding (autoregressively predicting the next token). The dominant source of memory overhead varies by task; for example, long chain-of-thought is more memory-intensive during decoding, whereas long-document QA incurs more overhead in pre-filling.

\item Methods utilize different notions of sparsity, for example, some measure attention sparsity~\citep{jiang2024minference}, while others focus on KV cache sizes~\citep{xiao2025duoattention}.

\item KV eviction is useful only if performance is preserved within tolerable limits, but methods often compete in terms of performance, even if no method preserves performance within these limits.

\end{enumerate}
We bridge these differences by introducing a new metric: the \textbf{KV footprint}, which equals the time-aggregated KV cache memory usage. 
It captures overhead from both pre-filling and decoding, and enables a comparison of different methods on equal footing.
We define the \textbf{critical KV footprint} as the minimum footprint that retains no less than $\mathcal F = 90\%$ of the performance relative to full attention.

We catalog existing approaches to KV eviction into several classes and analyze their critical footprint.
Most approaches have a high KV footprint in one or both stages of inference (Section~\ref{sub:approaches}).
For example, 
PyramidKV~\citep{cai2024pyramidkv},
a \emph{post-fill eviction} method that evicts KVs based on the attention pattern of the prompt,
incurs high peak memory and hence high KV footprint when encoding the input sequence. 
We introduce \textit{chunked eviction} in Section~\ref{sec:chunked_eviction}, which lowers the KV footprint substantially for such methods  by allowing the model to evict KVs earlier during pre-filling.

We then consider %
DuoAttention~\citep{xiao2025duoattention}, %
a promising approach that specializes attention heads into either local or global information.
This distinction is learned based on the hidden representation similarity between the two types in a synthetic recall task.
We introduce {\bf PruLong} in Section~\ref{sub:duo}, a new training method where the type of attention head is learned from diverse long-context pre-training data based on the next-token prediction loss, and the attention type is gradually discretized throughout training.
The improved objective and training data allow \ours{} to learn more accurate attention head assignments, helping it deliver more performance at smaller KV footprints.

We evaluate these methods on diverse long-context task categories, including recall, RAG, and many-shot in-context learning from HELMET~\citep{yen2025helmet}, as well as procedural generation and reasoning tasks from LongProc~\citep{ye2025longprocbenchmarkinglongcontextlanguage}.
Our evaluation provides a stress-test of KV eviction methods with input sequences of up to 110K tokens and long  generations of 10K tokens.
In our evaluation, none of the methods can consistently attain the best critical footprint across all tasks.
PruLong and chunked eviction often outperform prior methods by lowering the critical KV footprint by at least 10\%. PruLong is notably strong on recall tasks, but we also observe that it is sensitive to the pre-filling chunk size.
Finally, we study whether applying PruLong to a base long-context model (before instruction tuning) can temper this issue, but we notice a similar sensitivity due to a mismatch between training and inference. In summary, our contributions are:

\begin{itemize}[topsep=0pt,parsep=0pt,partopsep=0pt, leftmargin=2em]
    \item We introduce a conceptual framework that unifies existing KV eviction methods, and compares them fairly based on their \emph{critical KV footprint} (\autoref{sec:setup}). %
    
    \item We categorize existing methods in Section~\ref{sec:setup} and carefully consider how they may be modified to achieve lower critical KV footprints via \emph{chunked eviction} (\autoref{sec:chunked_eviction}). We then design \emph{\ours{}}, which identifies local and global attention heads via end-to-end training with the next-token prediction loss (\autoref{sub:duo}).
    
    \item In our challenging long-context evaluations, we show that PruLong outperforms DuoAttention in the majority of tasks and reduces the critical KV footprint by an additional 12\% in recall tasks. 
    However, chunked PyramidKV eviction is the most effective method for retaining performance in ICL and RAG settings
    (\autoref{sec:experiments}).
         
\end{itemize}
To encourage further research into robust KV eviction methods, we release all our code and artifacts.

%% file: sections/02_background.tex
\section{A Unified Framework for KV Cache Eviction}
\label{sec:setup}

\subsection{Measuring the Critical KV Footprint}

Given a prompt $\mathbf{x}_{1:n}$ with $n$ tokens, a transformer-based language model conventionally generates a response $\mathbf{x}_{n+1:n+m}$ in two stages:
\begin{enumerate}[topsep=0pt,parsep=0pt,partopsep=0pt, leftmargin=2em]
    \item {\bf Pre-filling}: The entire prompt $\mathbf{x}_{1:n}$ is processed in one forward pass.
    The key-value states $\{\mathbf{k}^{h}_{1:n}, \mathbf{v}^{h}_{1:n}\}$ for each attention head $h$ are stored in a \emph{KV cache}. 
    Here $\mathbf{k}^{h}_{i},\mathbf{v}^{h}_{i}\in \mathbb{R}^{d}$ where $d$ is the head dimension. 

    \item {\bf Decoding}: Tokens $\mathbf{x}_{n+1}, \mathbf{x}_{n+2}, \ldots \mathbf{x}_{n+m}$ are decoded one-by-one, reading and updating the KV cache each time. 

\end{enumerate}

The memory consumption of the KV cache grows linearly with increasing prompt and generation lengths, 
and researchers have proposed many methods to address this overhead.
Broadly, these methods work by sparsifying the attention patterns and hence allowing certain KV entries to be evicted. 
Still, they are tailored to different stages in the inference pipeline---some discard KV entries after the pre-fill stage (e.g.~\citet{cai2024pyramidkv}), while others also trim the KV cache during pre-fill (e.g.~\citet{xiao2025duoattention}). 
This makes a fair and holistic comparison between methods difficult. 
We first discuss why the common metric of KV cache size fails to capture the model’s usefulness in real-world applications.

In practice, it is expensive  to conduct only a single pass of pre-fill for long-context inference. %
\emph{Chunked pre-filling}---splitting the input sequence into chunks and processing it in multiple forward passes---is increasingly standard practice for long input sequences\footnote{For instance, SGLang~\citep{zheng2024sglang} pre-fills long inputs in chunks of 8192 tokens by default.}.
This typically reduces the peak GPU memory associated with long inputs\footnote{
Both the KV states and the hidden states and query layer contribute towards the peak memory. With grouped-query attention~\citep{ainslie2023gqa} or multi-latent attention~\citep{liu2024deepseek}, the latter two can have much higher footprints than the KV states for a layer, and pre-filling smaller chunks reduces the size of hidden states and query vectors present at any given step in memory.} and enables decoding for shorter prompts to co-occur with additional chunks of longer prompts~\citep{agrawal2023sarathi}.
While beyond the scope of this paper, scenarios like multi-turn conversations or interleaved tool calling also necessitate multiple decoding and pre-fill stages, requiring a holistic approach to measuring KV footprint. Speculative decoding~\citep{leviathan2022fast, sadhukhan2025magicdec} further blurs the lines between the pre-fill and decoding stages, by making decoding more compute-bound.

When considering inference with multiple forward passes during both pre-filling and, of course, decoding, the ``KV footprint'' should account for memory usage over time. For example, it should reflect if we already evicted KVs in chunked pre-filling before pre-filling is completed.
The exact inference is characterized by input and output lengths, as well as implementation details that vary across methods. 
In the absence of a metric that can capture all these nuances, we propose an idealized metric that
(1) tracks KV cache memory usage throughout pre-filling and decoding, and
(2) accounts for the lifespan of each KV entry,
enabling apples-to-apples comparisons across methods.

\begin{minipage}{0.55\linewidth}

We examine the methods' attention patterns (Figure~\ref{fig:definition}) and classify each KV entry as \emph{active} (used in the current step), \emph{inactive} (stored but not used at the current step), or \emph{evicted} (not used at any future step and removed from memory).
We define the \textbf{KV footprint} as the number of attention entries that have not been \emph{evicted}, aggregated across all timesteps.
This number is normalized to full causal attention.
For example, in \autoref{fig:definition}, the 
KV footprint is $26/36 = 72.2\%$.
An ideal method minimizes the footprint by evicting KVs \emph{as early} as possible.

\end{minipage}\hfill
\begin{minipage}{0.42\linewidth}
    \centering
    \includegraphics[width=\linewidth]{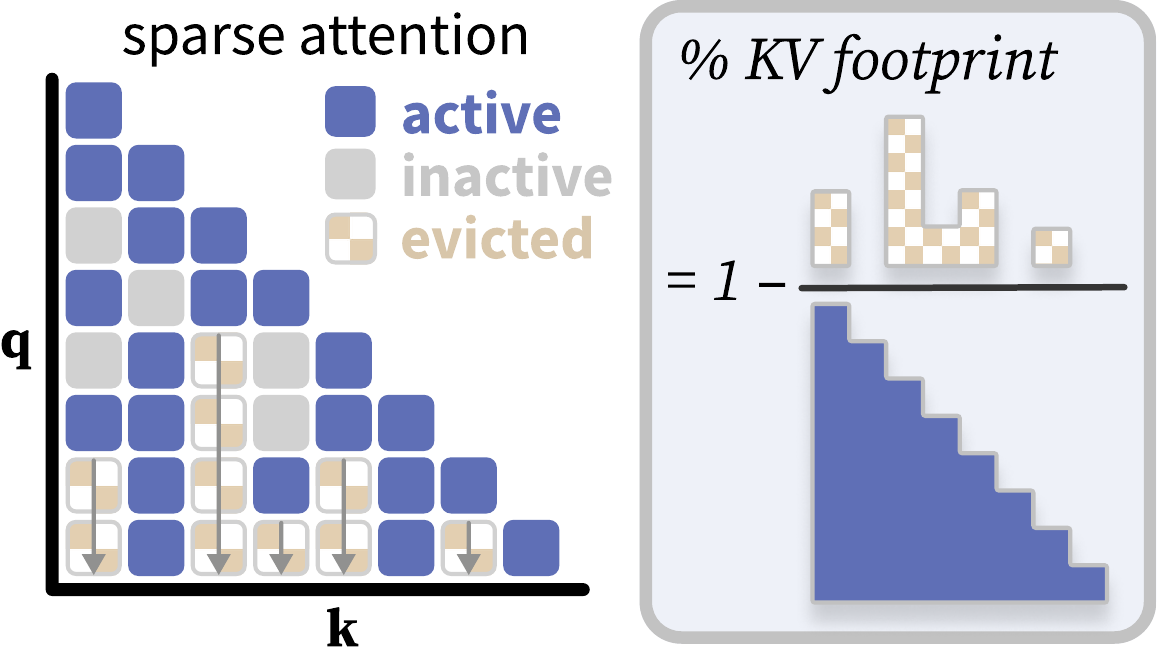}
    \captionof{figure}{The KV footprint. 
    }
    \label{fig:definition}
\end{minipage}

In \autoref{app:kv_metrics}, we consider an alternative metric that tracks peak KV occupancy in the attention matrix, and both metrics yield similar takeaways in our experiments. We also discuss how our methods relate to real performance metrics such as total token throughput and GPU memory utilization. 
We find that in many cases the KV footprint correlates well with throughput, yet the precise rankings depend on implementation details beyond KV evictions---and practical efficiency varies widely for the different implementation frameworks across methods.

\paragraph{The critical KV footprint.}
Prior work often reports task performance at a fixed sparsity level~\citep{li2025scbench}, but we argue a more meaningful metric is the sparsity achievable while preserving most of the original performance.
We define the \emph{critical KV footprint} as the smallest footprint at which a method retains a fraction $\mathcal{F}$ of the full-attention performance on a long-context task.
This paper uses $\mathcal F = 90\%$.
Below this threshold, the performance drop is likely too severe for the method to remain attractive in practice.
Our approach differs from SCBench~\citep{li2025scbench}, who propose separate performance metrics for each generation scenario (pre-fill, decoding, and multi-turn) and report their main results at a fixed sparsity setting for each method.

\subsection{Existing Approaches to Efficient Long-Context Inference}
\label{sub:approaches}

We survey efficient long-context methods and discuss how they fit into our framework of KV footprint. \autoref{tab:comparemethods} gives an overview over prominent approaches, showing how methods make different trade-offs and use different notions of sparsity.

\begin{figure}[t]
    \centering
    \includegraphics[width=\linewidth]{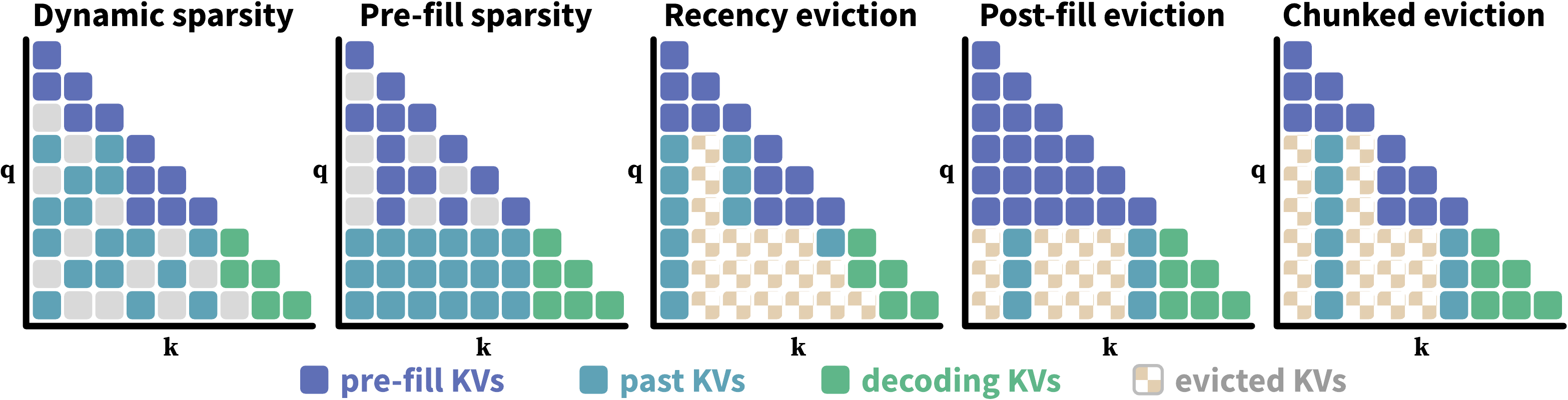} 
    \caption{Overview of relevant methods for efficient inference discussed in Section~\ref{sub:approaches} and Section~\ref{sec:chunked_eviction}.
    The example visualizes the generation of 3 output tokens after two steps of chunked pre-filling, except for pre-fill sparsity and post-fill eviction, which require a single pre-fill.
    }
    \label{fig:methods}
\end{figure}

\input{tables/methods}

\paragraph{Dynamic and pre-fill sparsity.}
Native Sparse Attention~\citep{yuan2025native}, MoBA~\citep{lu2025moba}, QUEST~\citep{tang2024quest}, and TokenButler~\citep{akhauri2025token} treat the KV cache as a two-level hierarchy, where only relevant attention blocks are loaded from high-bandwidth memory (HBM) to on-chip SRAM for processing.
Techniques like MInference~\citep{jiang2024minference} or FTP~\citep{xu2025ftp} use dynamic sparse attention to approximate the full attention specifically during the pre-filling phase.
Dynamic sparsity approaches lead to more \emph{inactive} KVs and can improve throughput, but they do not reduce the KV memory, and therefore these methods are orthogonal to our focus.

\paragraph{Recency eviction.}
\citet{xiao2024efficient} identify streaming attention heads, which only attend to a local sliding window and a set of initial ``sink tokens''. Evicting distant KV entries reduces the KV footprint substantially (Figure~\ref{fig:methods}), as the size of the KV cache remains fixed despite increasing context length, and this can be applied during both pre-filling and decoding.
However, recency eviction may ``forget'' relevant distant context, motivating
DuoAttention~\citep{xiao2025duoattention} and MoA~\citep{fu2024moa} to convert
only a subset of attention heads to streaming heads. As promising candidates to KV cache compression, we discuss these methods in more detail in~\autoref{sub:duo}.

\paragraph{Post-fill eviction.}
We use the term \emph{post-fill eviction} to refer to methods that drop tokens from the KV cache after the prefill phase ends.
These methods rely on heuristics typically based on attention scores \citep{zhang2023h2o, ge2024model, li2024snapkv, cai2024pyramidkv, chen2024nacl, wang2025llms} to identify the most important KVs in the context.
These methods can heavily trim the KVs after the pre-fill and reduce KV memory during decoding. 
However, in inference scenarios with long prompts and short generations, 
post-fill eviction can only achieve a limited reduction of the KV footprint, 
as all KV entries are kept in memory during pre-fill, which also leads to a substantial peak memory before eviction.
We adapt these methods for the chunked pre-filling setting in \autoref{sec:chunked_eviction}.

\paragraph{Orthogonal techniques.}
Quantization~\citep{lin2023awq, xiao2023smoothquant, frantar2023gptq} saves memory by reducing the precision rather than the cardinality of the KV cache,
and could be combined with any method considered in this paper.
Another direction is to design memory-efficient architectures before pre-training a new language model.
This may involve re-using the KV states across queries~\citep{ainslie2023gqa} or layers~\citep{sun2024you, qiao2024swiftkv}, reducing the key-value dimensionality~\citep{deepseekai2024deepseekv2, zhang2025tensor}, or interleaving global and local attention layers~\citep{rajput2024inference, gemmateam2025gemma3}.
Other approaches replace softmax attention with recurrent layers~\citep{peng2023rwkv}, linear attention~\citep{sun2023retentive, katharopoulos2020linear} or state-space layers~\citep{gu2022efficiently, gu2024mamba, poli2023hyena, wang2024the}.
These methods are orthogonal to KV eviction, and we leave their analysis to future work.

%% file: tables/methods.tex
\newcommand{\cmark}{\ding{51}}%
\newcommand{\xmark}{\ding{55}}%

\begin{table}[t]
    \footnotesize
    \caption{Comparison of prior works for efficient long-context inference. Methods measure sparsity in varied ways and focus on different stages of the generation process, and not all methods are relevant to our goal of reducing the KV cache footprint.}
    \centering
    \begin{tabular}{lll}
        \toprule
        \textbf{Approach} & \textbf{Sparsity Measure} & \textbf{Pros/Cons} \\
        \midrule 
        \textbf{Dynamic Sparsity} & \% inactive & \textcolor{Green!70!black}{\checkmark Fewer memory accesses} \\
        $\;\;$NSA \citep{yuan2025native} & attention weights & \textcolor{Green!70!black}{\checkmark Increased throughput} \\
        $\;\;$MoBA \citep{lu2025moba} &  & \textcolor{red!90!black}{\xmark\ No reduction in KV footprint} \\
        $\;\;$TokenButler \citep{akhauri2025token} & & \\
        \midrule
        \textbf{Pre-fill Sparsity} & \% inactive & \textcolor{Green!70!black}{\checkmark Speeds up pre-fill}  \\
        $\;\;$MInference \citep{jiang2024minference} & attention weights  & \textcolor{red!90!black}{\xmark\ No reduction in KV footprint}\\
        $\;\;$FTP~\citep{xu2025ftp} & (pre-fill only) & \textcolor{red!90!black}{\xmark\ No impact during decoding }\\
        \midrule
        \textbf{Recency Eviction} & \% local heads; & \textcolor{Green!70!black}{\checkmark Reduces KV footprint} \\
        $\;\;$StreamingLLM \citep{xiao2024efficient} & local window size & \textcolor{Green!70!black}{\checkmark Compatible with chunked pre-filling} \\
        $\;\;$DuoAttention \citep{xiao2025duoattention} &   & \textcolor{red!90!black}  {\xmark\ Fixed, context-invariant sparsity}  \\
        $\;\;$MoA \citep{fu2024moa} & & \\
        \midrule
        \textbf{Post-fill Eviction} & \% evicted KVs & \textcolor{Green!70!black}{\checkmark Reduced KV memory during decoding} \\
        $\;\;$H$_2$O \citep{zhang2023h2o} & after pre-fill & \textcolor{red!90!black}{\xmark\ High peak memory during pre-fill} \\
        $\;\;$FastGen \citep{ge2024model} & & \textcolor{red!90!black}{\xmark\ Only evaluated with single pre-fill} \\
        $\;\;$SnapKV \citep{li2024snapkv} & & \textcolor{red!90!black} {\xmark\ No KV reduction for long generations } \\
        $\;\;$PyramidKV \citep{cai2024pyramidkv} & & \\
        \bottomrule
    \end{tabular}
    \label{tab:comparemethods}
\end{table}

%% file: sections/03_method.tex
\section{Chunked Eviction: Making Post-Fill Eviction Pre-Fill Friendly}
\label{sec:chunked_eviction}

\autoref{fig:methods} illustrates that post-fill eviction has a limited scope for reducing the overall KV footprint for long input sequences, since all KVs are retained during the pre-fill and temporarily lead to a high peak memory.
Chunked pre-filling presents an opportunity to reduce the KV footprint by evicting KV entries after each chunk is processed. We visualize this kind of \emph{chunked eviction} in Figure~\ref{fig:methods}.
This is in contrast to storing all KVs until the end of chunked pre-filling, which would be functionally equivalent to post-fill eviction but exemplifies why this incurs a large KV footprint in our framework.

We adapt two popular post-fill eviction methods, SnapKV~\citep{li2024snapkv} and PyramidKV~\citep{cai2024pyramidkv} to the chunked pre-filling setting. %
Both methods use the total attention received by a KV entry from the last $k$ input tokens (where $k = 64$) as a proxy for its importance during decoding, and evict all but the highest-scoring KVs (after smoothing the attention with a moving average).
The two methods primarily differ in how they allocate the KV budget across layers. 
SnapKV selects an equal number of KVs in each layer, whereas PyramidKV uses a successively smaller budget for later layers.
We focus on PyramidKV in the main text to avoid clutter since it generally performs better, but present results with SnapKV in~\autoref{app:results}.

We investigate two ways of making these methods compatible with chunked-prefilling:
\begin{enumerate}[topsep=0pt,parsep=0pt,partopsep=0pt, leftmargin=2em]
    \item \textbf{Naive chunked eviction} simply applies the eviction heuristic after the forward pass of each chunk, using the last $k$ of the chunk to decide which KV entries to evict. 
    \item \textbf{Patched chunked eviction} respects that token importance should be computed based  on the last $k$ tokens of the prompt. In practice, we append these query tokens to the end of each chunk before computing the forward pass, but discard their KVs unless we process the final chunk.
\end{enumerate}

Besides enabling chunked eviction, we notice that both PyramidKV and SnapKV may actually {\it increase} the KV footprint of LMs using grouped-query-attention (e.g., all Llama models), as the KV entries are replicated across each query group to select separate entries based on the attention scores of each individual query head. We address this oversight by selecting a single set of KV entries for each query group based on the total attention scores across queries---in small-scale experiments, this improves performance while reducing memory usage by a factor of 8x for Llama-3.1-8B-Instruct.

We note that there are a few eviction methods~\citep{devoto2024simple, huang2025locret} which compute the importance of a KV entry only using local representations (e.g., evict based on the $L_2$ norm of the key vector), which works out-of-the-box with chunked pre-filling.

\section{\ours{}: An End-to-End Approach for Specializing Attention Heads}
\label{sub:duo}

We discussed in Section~\ref{sub:approaches} how evicting ``stale'' KVs may substantially reduce the KV footprint, but risks losing important past information.
This has motivated follow-up work to understand {\it which} attention heads focus on global vs. local context, and only evict KVs in local attention heads.

DuoAttention~\citep{xiao2025duoattention} categorizes attention heads into two types: \emph{retrieval heads}, which recall relevant information from the entire context and \emph{streaming heads}~\citep{xiao2024efficient}, which only attend to recent tokens and a small set of ``sink'' tokens at the beginning of the input sequence.
DuoAttention learns the attention head type by expressing the attention mechanism as a superposition of streaming and full attention, parametrized by
\begin{equation}
\text{Attn}_{i,j}(\mathbf{Q}, \mathbf{K}, \mathbf{V}) = z_{i,j} \cdot \text{Attn}_\text{full}(\mathbf{Q}, \mathbf{K}, \mathbf{V}) + (1 - z_{i,j}) \cdot \text{Attn}_\text{streaming}(\mathbf{Q}, \mathbf{K}, \mathbf{V})
\label{eq:duo_attention}
\end{equation}
where $i$ and $j$ run over the $L$ layers and $H$ attention heads of the transformer, respectively.
The masks $z_{i,j}$ are trained with an $L_2$ reconstruction loss between the final hidden states of the original and interpolated models, and the masks $z$ are encouraged to be sparse via $L_1$ regularization.
DuoAttention uses long-context training data which consists of synthetic needle-in-a-haystack tasks.
Upon convergence, a head sparsity of $s\%$ is obtained by setting the bottom $s\%$ of $z_{i,j}$ to $0$ and the rest to $1$.
MoA~\citep{fu2024moa} is another method that uses natural text, but is difficult to scale beyond sequences longer than 8K tokens, as it materializes the full attention matrix.

While DuoAttention shows strong empirical performance, we identify several ways to push its critical KV footprint even lower.
We combine these insights to design \textbf{\ours{}}, an end-to-end method for KV eviction.
PruLong classifies attention heads into one of the two roles like Duo, but innovates upon the training objective, parametrization, and training data. 
We now describe each in turn.

\begin{enumerate}[topsep=0pt,parsep=0pt,partopsep=0pt, leftmargin=2em]
\item \textbf {Next-token prediction loss.}
\ours{} directly minimizes the next-token prediction loss of the hybrid attention model, rather than the reconstruction error of the last hidden state, aligning better with how these models are used in text generations.

\item \textbf{Optimizing discrete masks over attention types.}
DuoAttention learns a continuous gating variable $z_{i,j} \in [0, 1]$, which is easy to optimize, but does not reflect that $z_{i,j}$ will be rounded to $0$ or $1$ during inference, therefore introducing a train-test gap.
\ours{} treats $z_{i,j}$ as binary masks drawn from a Bernoulli distribution parameterized by $\pi_{i,j}$, and enable end-to-end optimization via established approaches from the pruning literature~\citep{louizos2018learning}---reparameterizing Bernoulli distributions as \textit{hard concrete} random variables.
The final objective is as follows:
\begin{equation}
    \max_{\lambda_1, \lambda_2}  \min_{\mathbf{\pi}} \mathop{\mathbb{E}}\limits_{\substack{\mathbf{x} \sim D\\\mathbf{z} \sim \text{Bern}(\mathbf{\pi})}} \underbrace{\left[\frac{1}{N}\sum_{n=0}^{N-1}\log p_\theta(\mathbf{x}_{n+1}|\mathbf{x}_{:n}; \mathbf{z})\right]}_{\mathcal L_\text{next-token}} + \underbrace{\lambda_1 \left(s(\mathbf{\pi})-t\right) + \lambda_2(s(\mathbf{\pi})-t)^2}_{\mathcal{L}_\text{reg}},
    \label{eq:totalobj}
\end{equation}
where $\mathcal{L}_\text{reg}$ constraints the overall sparsity of the masks $s(\pi)$ towards a target value $t$.
This is enabled via min-max optimization, where $\lambda_1$ and $\lambda_2$ are trainable Lagrange parameters optimized via gradient \emph{ascent}.
We provide more details in Appendix~\ref{ap:objective}.

\item \textbf{\ours{} leverages natural long-context data.}
DuoAttention’s synthetic training data only requires simple long-range recall, whereas real-world application may demand more complex abilities.
\ours{} is trained on natural long-context pre-training data (such as code repositories and books) by~\citet{gao2025how}, containing diverse long-range dependencies.

\end{enumerate}

%% file: sections/04_experiments.tex
\vspace*{-0.6em}
\section{Experiments}
\label{sec:experiments}
\vspace*{-0.5em}

\subsection{Evaluation Setting}
\label{sec:eval}

\paragraph{Diverse and challenging tasks.} 
Our evaluation consists of tasks from HELMET~\citep{yen2025helmet} (long inputs $\rightarrow$ short outputs) and LongProc~\citep{ye2025longprocbenchmarkinglongcontextlanguage} (short/long inputs $\rightarrow$ long outputs) on which Llama-3.1-8B-Instruct (the model we use for evaluation) achieves non-trivial performance.
We evaluate the HELMET tasks at the 128K context setting to stress-test KV reduction methods with information-rich contexts.
Overall, we evaluate on 21 datasets report average performance across 8 task categories, which cover various long-context applications (RAG, reranking, summarization) and capabilities (recall, reasoning---in travel planning, ignoring distractions---in RAG, and in-context learning). HELMET also covers recall tasks sourced from RULER~\citep{hsieh2024ruler}, as well as QA and summarization tasks from $\infty$ Bench~\citep{zhang-etal-2024-bench}.
A detailed description of the task setup, datasets, and input and output lengths is provided in Appendix~\ref{app:experiments}.

\paragraph{Methods and hyperparameters.}
We use Llama-3.1-8B-Instruct~\citep{dubey2024llama} as a capable long-context language model.
To obtain a wide range of KV footprints, we evaluate each method with a grid of hyperparameters.
We run DuoAttention and PruLong with head sparsities ranging from 10\% to 90\%, but fix the local window size to be 1024 tokens and use 128 attention sink tokens---this has a comparatively small impact on the KV footprint given the long context lengths of our evaluation tasks. We were not able to run MoA~\citep{fu2024moa} since its training method did not scale to sequences longer than 8K tokens.
For chunked eviction, our primary focus is PyramidKV~\citep{cai2024pyramidkv}, as it performs slightly better than SnapKV~\citep{li2024snapkv} (see comparison in~\autoref{app:results}), and far better than key magnitude-based eviction by~\citep{devoto2024simple} in our setting.
We use the recommended setting of using the last 64 input tokens to compute KV importance, and evict $p\%$ of input KVs at each pre-filling step, where $p\%$ also ranges from 10\% to 90\%. 
Unless otherwise noted, we evaluate methods with a pre-filling chunk size of 32K.

\subsection{How many KVs Are Needed for Long-Context Abilities?}
\label{sec:results}

\input{tables/main_results}

\begin{figure}[t]
    \centering
    \includegraphics[width=1.0\textwidth]{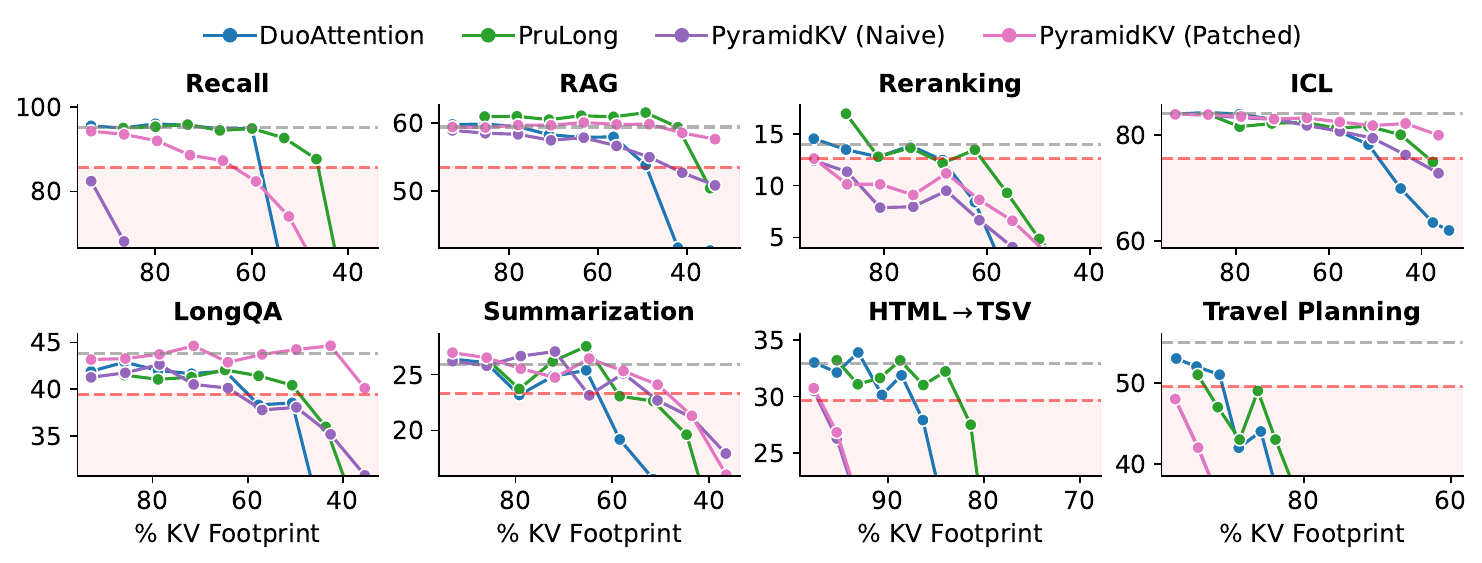}
    \caption{Performance vs. KV footprint for the baselines and  \ours{} (green). The gray dashed line denotes the original model's performance, and the red one represents $90\%$ of model performance.}
    \label{fig:sparsity_trends}
\end{figure}

\autoref{fig:sparsity_trends} visualizes the efficacy of the methods with varying KV footprint. \autoref{tab:min_kv_footprint} summarizes the critical KV footprint, i.e., the smallest value for which task performance remains within 90\% of the performance with a full KV cache.
We report results with other metrics of KV usage in \autoref{app:experiments}. 

\paragraph{PruLong reliably reduces KV footprint for recall tasks.}
Recall is a good stress-test of long-context modeling ability, as it directly evaluates the ability to pick out relevant information from long ago, without confounding factors like the model's ability to reason about the retrieved information.
We see that DuoAttention and \ours{} excel on this task in comparison to more heuristic eviction methods.
In particular, PruLong reduces the critical KV footprint by roughly 12  points compared to its predecessor.
PruLong also exhibits the lowest critical footprint on re-ranking, reordering retrieved passages from MS MARCO~\citep{bajaj2016ms}, and HTML $\to$ TSV, a structured prediction task, and improves upon DuoAttention on all tasks except Travel Planning.
Even though patched PyramidKV achieves substantially lower critical KV footprints on LongQA and ICL, it is not clear whether this would be truly reliable in practice given its recall performance.
The difference between the two methods is especially pronounced when targeting low KV footprints.

\paragraph{Patching is important for chunked eviction.}
Chunked eviction allows PyramidKV to reduce KV footprints meaningfully; with a single pre-fill the minimum KV footprint if bounded by 0.1-5\% depending on the task as reported in ~\autoref{app:results}.
In \autoref{fig:sparsity_trends}, we observe that patching is important for retaining reliable performance in tasks---allowing PyramidKV to come out as the best method on ICL, RAG, LongQA, and Summarization.
While PyramidKV still lags behind Duo on recall tasks, patching reduces the critical KV footprint by 30\% compared to naive eviction.

\begin{figure}[t]
    \centering
    \hspace{2em}
    \includegraphics[scale=0.5]{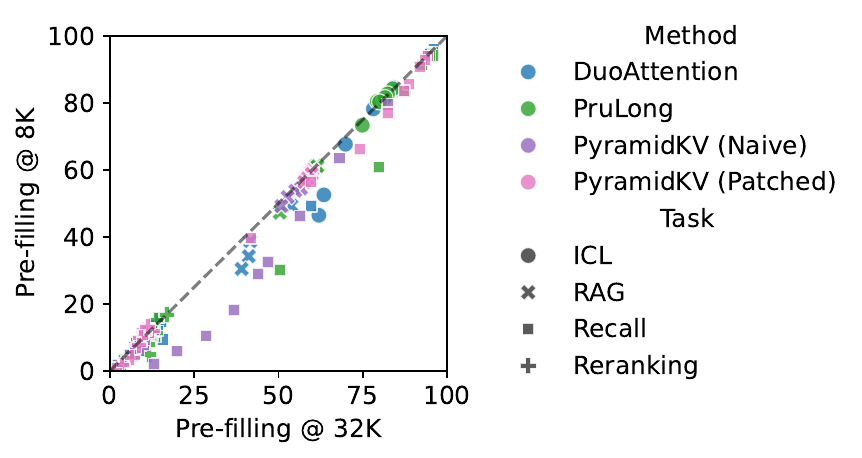}%
    \hfill%
    \includegraphics[scale=0.5]{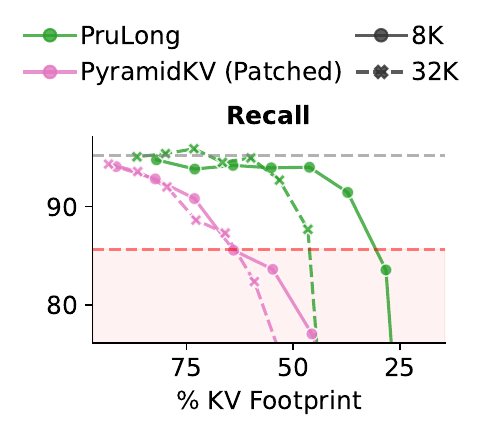}%
    \hspace{2em}
    \caption{Left: Pre-filling with a smaller chunk size (8K tokens vs. 32K tokens) tends to deteriorate performance on the same task. The axes track the performance on the four tasks at the specified pre-fill length. Right: Pre-filling with smaller chunks (8K tokens) achieves smaller KV footprints and lies on the pareto frontier, particularly when aiming for high KV reduction.}
    \label{fig:chunk_size}
    \vspace{-1em}
\end{figure}

\paragraph{Sensitivity to pre-fill chunk size.} 
We observe in Figure~\ref{fig:chunk_size}, that suprisingly, both DuoAttention and PruLong are more susceptible to performance loss (up to 20\%) when reducing the chunk size from 32K to 8K than patched chunked eviction. However, smaller chunk sizes dominate the Pareto frontier as they evict KVs more frequently. In terms of reliability, we also note that no method achieves any meaningful reduction in KV footprint on the reasoning-heavy Travel Planning task. 

\input{tables/prulong_ablations}

\subsection{Explaining the Strength of PruLong}

\begin{figure}
    \centering
    \includegraphics[width=1.0\linewidth]{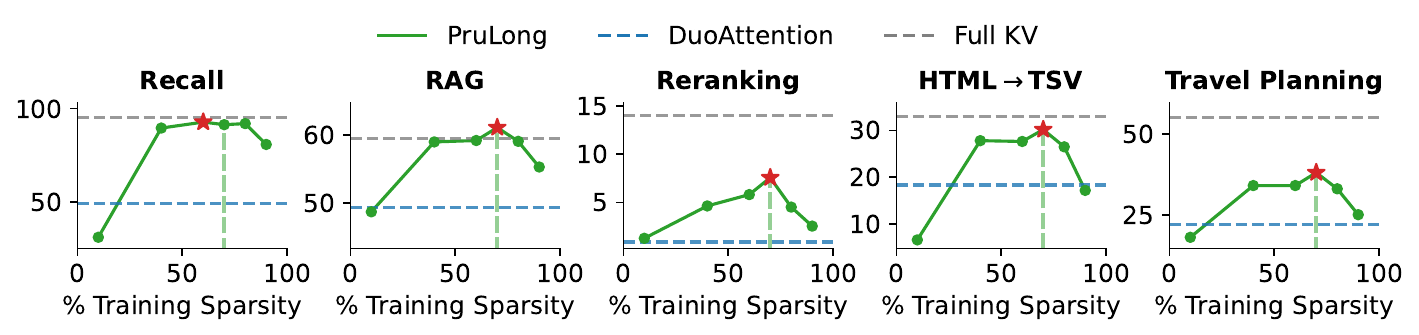}
    \caption{We investigate how masks trained with different sparsities perform when evaluated at $70\%$ sparsity. Regularization with target sparsity leads to better performance across the board, although PruLong outperforms DuoAttention at other sparsities as well.}
    \label{fig:training_sparsity}
\end{figure}

We investigates which factors explain the superior performance of PruLong compared to DuoAttention. Here, we evaluate both methods at a fixed KV footprint corresponding to a 70\%  of attention heads used as streaming heads and 8K chunked pre-filling.
\begin{enumerate}[topsep=0pt,parsep=0pt,partopsep=0pt, leftmargin=2em]
    \item \textbf{\ours{} works better with natural long data.} 
    DuoAttention creates passkey retrieval data to learn the streaming heads. To capture more diverse long-context operations, we perform language modeling on broad long-context pre-training data sourced from \citet{gao2025how}. In Table~\ref{tab:prulong_ablations}, we show that unlike our method, DuoAttention does not learn well with noisy pre-training data, even when increasing the training budget by 4 times.
    We also confirm that our method favors pre-training data over Context Synthesis---a well-performing long-context instruction tuning dataset~\citep{zhu2025generalizing}. Note that we don't update regular model weights during training, as this would degrade the instruction-following abilities of the model.  

    \item \textbf{Precise regularization.} Our objective allows us to train a model against a specific target sparsity used for evaluation. Figure~\ref{fig:training_sparsity} reveals the outcome of evaluating models trained with different target sparsities at a 70\% sparsity. We note that the correctly regularized model (green dashed line) achieves the highest performance (red marker) in many diverse task categories.
\end{enumerate}

\subsection{At What Training Stage Should We Use \ours{}?}
\label{sec:stage}

Instead of applying PruLong to the final instruction-tuned model, one could learn the attention head types for the long-context base model and keep them fixed during the instruction tuning stage.

\begin{minipage}{0.65\linewidth}
    Since the supervised fine-tuning (SFT) data of Llama-3.1-8B-Instruct is not public, 
    we use ProLong-8B~\citep{gao2025how}---based on Llama-3-8B but with known long-context data and SFT recipe---we provide details in~\autoref{app:sft}.
    We run experiments to compare whether PruLong should be applied before or after SFT. Since both the training data and the language modeling objective matches the training of the ProLong base model, we explore unfreezing the weights during the PruLong training process. In \autoref{fig:pretrain_prulong}, we observe an interesting trend where updating the weights of the model leads to the best Recall at a chunk pre-filling size of 128 tokens---corresponding to the sliding window block size during training---%
    but deteriorates at greater chunk sizes. 
    We hypothesize  that updating the model weights allows the model to specialize to a fixed attention window during training. However, this makes it sensitive to the distribution shift when changing the attention window via the pre-filling chunk size during inference.

\end{minipage}\hfill
\begin{minipage}{0.33\linewidth}
    \centering
    \includegraphics[width=\linewidth]{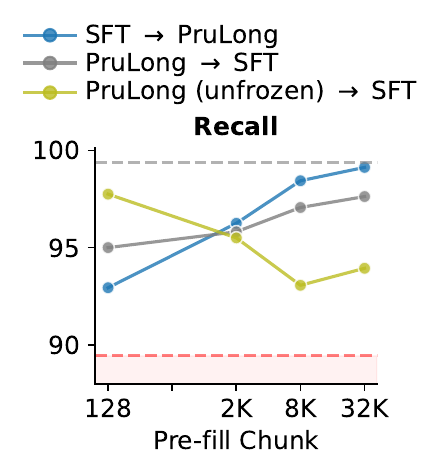}
    \captionof{figure}{PruLong applied to different training stages. Evaluated at a context length of 128K.}
    \label{fig:pretrain_prulong}
    \vspace{0.2em}
\end{minipage}

%% file: tables/main_results.tex
\begin{table}[t]
    \centering
    \small
    \setlength{\tabcolsep}{5pt} 
    \caption{The minimum effective KV footprint for various methods on HELMET and LongProc. 
    Due to the high cost of running the evaluation, 
    we interpolate this metric by linear interpolation of the data points in \autoref{fig:sparsity_trends}.}
    \begin{tabular}{lrrrrrrrr}
        \toprule
        \multirow{2}{*}{\textbf{Method}} & \multicolumn{8}{c}{\textbf{\textit{Critical KV footprint (\%) $\downarrow$}}} \\
        \cmidrule(lr){2-9}
         & \textbf{Recall} & \textbf{RAG} & \textbf{Re-Rank} & \textbf{ICL} & \textbf{LongQA} & \textbf{Summ} & \textbf{HTML} & \textbf{Travel} \\
        \midrule
        DuoAttention      & 58.0   & 49.0    & 69.0   & 49.0   & 60.0   & 63.0   & 87.0   & \bf 91.0    \\
        $\rightarrow$ PruLong                           & \bf 46.0   & 37.0  & \bf 61.0   & 38.0   & 49.0   & 59.0   & \bf 83.0   & 93.0    \\
        \addlinespace
        PyramidKV (Naive)      & $>$93.0 & 44.0    & $>$94.0 & 42.0   & 62.0   & 53.0   & 97.0   & $>$98.0  \\
        $\rightarrow$ PyramidKV (Patched)  & 64.0   & \bf $<$34.0 & 94.0   & \bf $<$36.0 & \bf $<$35.0 & \bf 49.0   & 97.0   & $>$98.0  \\
        \bottomrule
    \end{tabular}
    \label{tab:min_kv_footprint}
\end{table}

%% file: tables/prulong_ablations.tex
\begin{table}[t]
    \centering
    \caption{We explore different training settings for the DuoAttention/\ours{} methods. We report task performance at a fixed sparsity of 70\% streaming heads, where the color corresponds to retaining 90\% of the base performance.
    The gray shading indicates the default setting of each method.}
    \label{tab:prulong_ablations}
    \footnotesize
    \begin{tabular}{lcccccc}
\toprule
&
{\bf Recall} & 
{\bf RAG} & 
{\bf Rerank} & 
{\bf ICL} & 
{\bf HTML} &
{\bf Travel} \\
\midrule
\bf Llama-3.1-8B-Instruct
  & {95.2}
  & {59.5}
  & {14.0}
  & {83.9}
  & {32.9}
  & {55.0} \\
\midrule
\bf DuoAttention \\
\rowcolor{black!5!white} \quad -- BookSum Passkey
  & \rbox{49.2}
  & \rbox{49.3}
  & \rbox{0.9}
  & \gbox{78.2}
  & \rbox{18.3}
  & \rbox{22.0} \\
\quad -- Pre-training Mix
  & \rbox{38.6}
  & \gbox{51.9}
  & \rbox{2.1}
  & \gbox{77.4}
  & \rbox{17.0}
  & \rbox{43.0} \\
$\quad \quad$(+ 4x steps)
  & \rbox{39.3}
  & \rbox{46.2}
  & \rbox{0.8}
  & \rbox{73.6}
  & \rbox{17.0}
  & \rbox{45.0} \\
\bf PruLong \\
\rowcolor{black!5!white}  \quad -- Pre-training Mix
  & \gbox{91.4}
  & \gbox{61.1}
  & \rbox{7.6}
  & \gbox{81.6}
  & \gbox{30.2}
  & \rbox{38.0} \\
\quad -- BookSum Passkey
  & \rbox{65.2}
  & \gbox{55.3}
  & \rbox{1.8}
  & \gbox{80.8}
  & \rbox{16.9}
  & \rbox{23.0} \\
\quad -- Context Synthesis
  & \rbox{21.6}
  & \rbox{46.2}
  & \rbox{0.9}
  & \rbox{71.2}
  & \rbox{3.4}
  & \rbox{17.0} \\
\bottomrule
\end{tabular}
\end{table}

%% file: sections/05_conclusion.tex
\vspace{-0.4em}
\section{Conclusions and Future Work}
\label{sec:conclusion}

In this paper, we presented a unified view of various KV-cache sparsification techniques through the lens of a unifying metric: the \emph{critical KV footprint}.
We then studied how we might achieve a lower footprint using two promising classes of KV eviction: post-fill eviction and recency eviction.
We adapted post-fill eviction methods to evict KVs during intermediate stages of pre-filling via \emph{chunked eviction}.
In the recency eviction class, we proposed a new KV eviction method that uses structured sparsity: \emph{PruLong}.
PruLong optimizes the next token prediction loss while leveraging tools from the pruning literature to learn attention head roles from unlabeled text, allowing it to integrate natively with model training. 
In empirical evaluation, we found that recency eviction generally achieved a lower critical footprint than post-fill eviction.
PruLong achieved a $10-15\%$ reduction in the critical KV footprint over the next best method in 3 out of 6 tasks.
An ablation study of PruLong revealed that it was effective both before and after instruction tuning.
We hope that our results inspire future work to take a holistic view of the KV-eviction problem and tackle it natively during model training.

\paragraph{Limitations and future work.}
Several promising directions for future work exist.
For one, rather than evict all but a fixed local window of tokens, one may apply pruning methods to make flexible decisions depending on the context.
None of the methods achieve strong results across all tasks; future work should make KV eviction methods robust to wider applications.
Since the KV footprint assumes an idealized model, it may not correlate perfectly with throughput or other hardware metrics; future work may look into designing metrics that holistically capture the generation process. 
Finally, due to computational constraints, our experiments only focus on a single model.

\section*{Acknowledgements}

We are thankful to Howard Yen and Xi Ye for their feedback on an earlier draft of the paper.
We would also like to thank the Princeton NLP group for helpful discussions and advice.
This work is gratefully supported by an NSF CAREER award (IIS-2239290) and a grant from Intel.

%% file: appendices/kv_metrics.tex
\section{Alternatives to KV footprint}
\label{app:kv_metrics}
In Section~\ref{sec:setup}, we defined the KV footprint as the number of un-evicted (i.e., active or inactive) KV entries aggregated across the query dimension.
Since we interpret the axis of time to run along the query dimension, the KV footprint is closely tied to the time aggregate of GPU memory utilization.
One could also consider, then, the \emph{peak KV}---which we define as the maximum number of un-evicted KV entries across all query indices (once again including inactive entries).
This metric is similarly related to the peak GPU memory utilization of the KV cache.
We normalize the peak KV to the sequence length and express the result as a percentage.

\begin{figure}[h]
    \centering
    \includegraphics[width=1.0\textwidth]{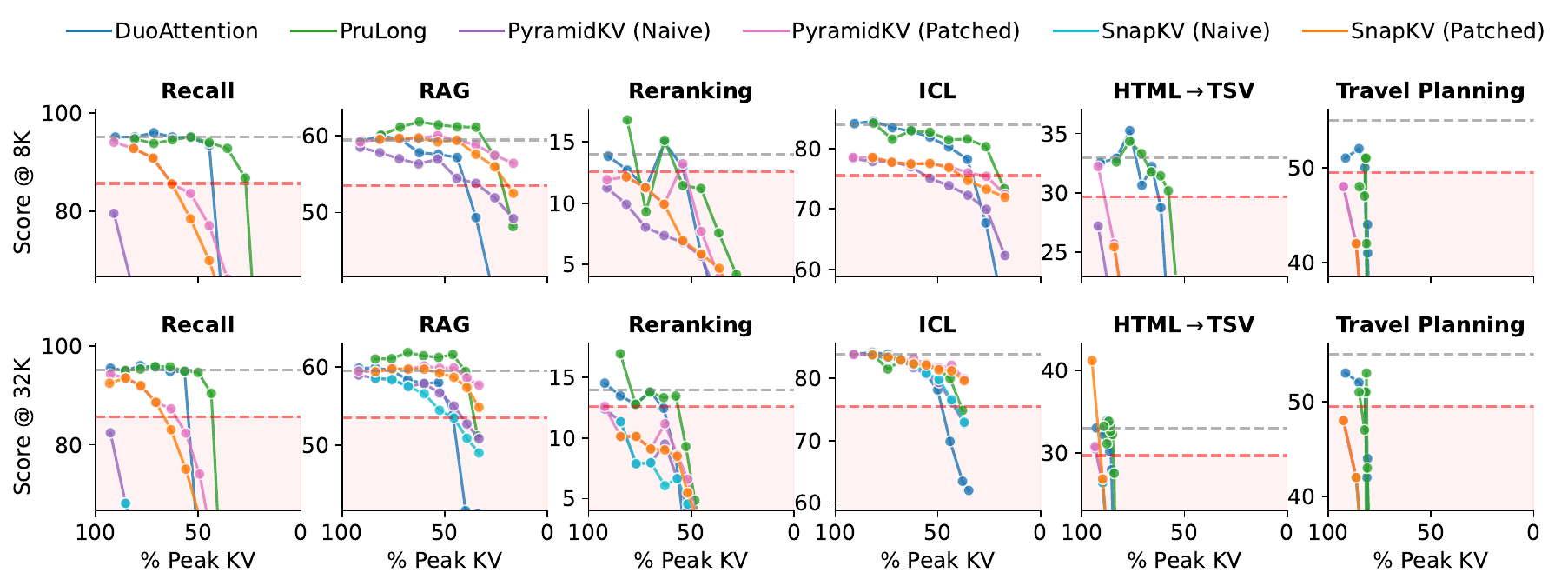}
    \caption{Performance vs. Peak KV for the baselines and \ours{}. The gray dashed line denotes the original model's performance, and the red one represents $90\%$ of model performance. We show results at both 8K and 32K prefilling chunk sizes.}
    \label{fig:peak_kv}
\end{figure}

We could then plot the score of the Llama-3.1-8B-Instruct model with different methods against the various peak KV values they achieve.
We do this in Figure~\ref{fig:peak_kv} for DuoAttention~\citep{xiao2025duoattention}, PruLong (ours), PyramidKV, and SnapKV.
For the latter two, we include results both with the naive version and with the patched version described in Section~\ref{sec:chunked_eviction}.
The results tell a story similar to Figure~\ref{fig:sparsity_trends}: on Recall, Rerank, HTML $\to$ TSV, and Travel Planning, PruLong and DuoAttention usually achieve lower peak KV percentages than PyramidKV and SnapKV.
In fact, PruLong's curves are strictly better than DuoAttention on all of the task groups.
Between PyramidKV and SnapKV, the former usually achieves a better score at the same footprint.
The addition of patching allows a further boost in the score, which in turn permits the method to achieve a local critical peak KV (which equals the peak KV at which the score drops below $90\%$ of the original model's).
We also note that PyramidKV and SnapKV worsen more at a pre-filling chunk size of 8K tokens compared to 32K, whereas DuoAttention and PruLong are more robust.

%% file: appendices/objective.tex
\section{The details of the pruning process}
\label{ap:objective}

This appendix describes the process of learning the mask parameters $\mathbf{z}$ in greater detail.
We reproduce below the objective used by \ours{}:

\begin{equation}
    \max_{\lambda_1, \lambda_2}  \min_{\mathbf{\pi}}  \mathop{\mathbb{E}}\limits_{\substack{\mathbf{x} \sim D\\\mathbf{z} \sim \text{Bern}(\mathbf{\pi})}} \underbrace{\left[\frac{1}{N}\sum_{n=0}^{N-1}\log p_\theta(\mathbf{x}_{n+1}|\mathbf{x}_{:n}; \mathbf{z})\right]}_{\mathcal L_\text{next-token}} + \underbrace{\lambda_1 \left(s(\mathbf{\pi})-t\right) + \lambda_2(s(\mathbf{\pi})-t)^2}_{\mathcal{L}_\text{lagrange}}
    \label{eq:totalobj2}
\end{equation}
The first term in Equation~\ref{eq:totalobj2} is the familiar next-token prediction loss over sequences $\mathbf{x}_{1:N}$ drawn from the training corpus $D$.
The second term is a Lagrangian penalty that forces a certain target sparsity $t$ on the masks $\mathbf{z}$.
We will now describe different aspects of the pruning process.

\paragraph{The hard-concrete reparametrization}
Objective~\ref{eq:totalobj2} parametrizes the masks $z_{i,j}$ as Bernoulli random variables with parameters $\pi_{i,j}$.
One may equivalently \emph{reparametrize} them in terms of the \emph{hard concrete distribution}~\citep{louizos2018learning}:
\begin{equation}
    \mathbf{u} \sim \text{Uniform}(0, 1)
    \label{eq:prune1}
\end{equation}
\begin{equation}
    \mathbf{s} = \sigma\left(\frac{1}{\tau} \cdot \log \frac{\mathbf{u}}{1 - \mathbf{u}} + \log \mathbf{\alpha}\right)
    \label{eq:prune2}
\end{equation}
\begin{equation}
    \tilde{\mathbf{g}} = l + \mathbf{g} \cdot (r - l)
    \label{eq:prune3}
\end{equation}
\begin{equation}
    \tilde{\mathbf{z}} = \min(1, \max(0, \tilde{\mathbf{g}}))
    \label{eq:prune4}
\end{equation}
Equation~\ref{eq:prune2} is a special case of the \emph{Gumbel} reparametrization~\citep{jang2017categorical}, also known as the Concrete distribution~\citep{maddison2017the}.
In the simple case of $\beta = 1$, it can be understood as a smooth relaxation of the indicator $\mathbb{I}(\log u_{i,j} - \log (1-u_{i,j}) + \log \alpha_{i,j} > 0)$, which itself is a Bernoulli random variable with success probability $\pi_{i,j} \triangleq \sigma(\log \alpha_{i,j})$.
The distribution rapidly converges to a discrete support $\{0,1\}$ as the temperature $\tau \to 0$. 
In practice, the uniform distribution is truncated at $(10^{-6}, 1 - 10^{-6})$, and the temperature $\tau$ is fixed at $\frac{3}{2}$.
Line~\ref{eq:prune3} then stretches this distribution to the interval $[-0.1, 1] \equiv [l, r]$, and the excess probability on either side is accumulated into a delta function at $0$ and $1$ (line~\ref{eq:prune4}).
This places a non-zero probability weight on the support ${0,1}$ to better represent the discrete nature of the modeled variables.
The hard concrete reparametrization allows us to re-express the expectation $\mathbb{E}_\mathbf{\pi}$ as $\mathbb{E}_\mathbf{u}$, which allows a gradient to be taken through the expectation using Monte Carlo sampling.
In this scheme, the parameters $\log \alpha$ are trainable and learned via gradient descent.

\paragraph{The Lagrange penalty} The expected $L_0$ sparsity $s = \mathbb{E} [||\mathbf{z}||_0]$ can be calculated in closed form as
$$s = 1 - \frac{1}{LH}\sum_{i,j} \mathbf{P}(z_{i,j} > 0 ) = 1 - \frac{1}{LH}\sum_{i,j} \sigma\left(\alpha_{i,j} - \log\frac{-l}{r}\right)$$
Then, the Lagrangian $\mathcal L_\text{lagrange}$ penalizes the deviation of $s$ from a desired sparsity $t$.
The parameters $\lambda_1$ and $\lambda_2$ are trainable and optimized with gradient \emph{ascent}, which forces the model to converge to $s = t$ to keep the objective low.
The target $t$ is warmed up over training from $0$ (which corresponds to the full model) to a desired value $t_\infty$ over the course of several (usually more than half of the total) training steps.

At the end of training, the top $k$ of the log alphas are marked up to $+\infty$ (corresponding to $z=1$), and the rest down to $-\infty$.
Any desired sparsity in $[0,1]$ may be achieved by choosing a suitable value of $k$, although best performance is obtained at sparsities near $t$ (Section~\ref{sec:experiments}).

%% file: appendices/experiments.tex
\section{Experimental Details}
\label{app:experiments}

\input{tables/tasks}

\subsection{Hyperparameters}
\begin{table}[h]
  \centering
  \caption{Hyperparameters used for \ours{}, SFT (in ablations), and for evaluation.}
  \label{tab:hyperparams1}
  \begin{tabular}{@{}ll@{}}
    \toprule
    \textbf{Hyperparameter}                       & \textbf{Value}  \\
    \midrule
    \multicolumn{2}{c}{\textit{\ours{}}}\\
    \midrule
    Batch size (tokens)                 & 1,048,576 \\
    Sequence length                     & 131,072\\
    Learning rate ($\log \alpha$)    & 1      \\
    Learning rate ($\lambda_1, \lambda_2$) & 1      \\
    Learning rate (model weights)          & frozen ($1\cdot10^{-5}$ in ablations) \\
    Training steps                 & 1,000   \\
    LR schedule                      & Linear warmup for first 10\% of steps \\
    & then linear decay to $1\%$ peak LR \\
    Adam $(\beta_1, \beta_2)$ & $(0.9, 0.95)$\\
    Initial target sparsity              & 0.0    \\
    Final target sparsity                & \{0.1,0.2,0.3,0.4,0.5,0.6,0.7,0.8,0.9,1.0\} \\
    Sparsity warmup steps                & 800    \\
    Local window size                    & 1,024   \\
    Sink size                            & 128    \\
    \midrule
    \multicolumn{2}{c}{\textit{SFT (in ablations)}}\\
    \midrule
    Batch size (tokens)                 & 4,194,304 \\
    Sequence length                     & 65,536\\
    Learning rate    & $2 \cdot 10^{-5}$      \\
    Training steps                 & 2500   \\
    LR schedule                      & Linear warmup for first 5\% of steps \\
    & then linear decay to $10\%$ peak LR \\
    Adam $(\beta_1, \beta_2)$ & $(0.9, 0.95)$\\
    Local window size                    & 1,024   \\
    Sink size                            & 128    \\
    \midrule
    \multicolumn{2}{c}{\textit{Evaluation}}\\
    \midrule
    Prefill chunk size                   & 32,768 (128 / 8,192 / 32,768 in ablations)  \\
    Evaluation sparsity (DuoAttention/\ours{})                & \{0.1,0.2,0.3,0.4,0.5,0.6,0.7,0.8,0.9,1.0\} \\
    Token retention sparsity (PyramidKV) & \{0.1,0.2,0.3,0.4,0.5,0.6,0.7,0.8,0.9,1.0\} \\
    Always retained window size (PyramidKV) & 64\\
    Patch amount (PyramidKV) & 64\\
    \bottomrule
  \end{tabular}
\end{table}

Unless otherwise stated, we use Llama-3.1-8B-Instruct~\citep{dubey2024llama} in our experiments.
The hyperparameters used for pruning, for the SFT in the ablations of Section~\ref{sec:experiments}, and during evaluation are listed in Table~\ref{tab:hyperparams1}.
Our default training data is derived from the stage-II continued pre-training mix by~\citet{gao2025how} (length 512K), which consists of a short and long data mixture component in a ratio of 40\% : 60\%.
We adjust the long data component to fit the context size of Llama-3.1-8B by truncating the 512K documents to length 128K, which also replace any 64K token documents in the long data component.

\subsection{Evaluation datasets}
\label{appsub:datasets}
We list the datasets that make up each task category of HELMET and LongProc in Table~\ref{tab:dataset_overview}.
Note that we evaluate across a wide range of long-context capabilities, including RAG, re-ranking, summarization, text extraction, and planning.
HELMET improves evaluation by providing in-context demonstrations and reliable metrics (e.g., model outputs judged by GPT-4o with respect to reference summary). We refer to the HELMET paper for details~\citep{yen2025helmet}.

\begin{table}[t]
\centering
\caption{Performance of different SFT data mixtures. We find that adding the Tulu-v3 SFT mixture~\citep{lambert2025tulu3} makes a difference especially on LongProc reasoning tasks such as HTML (HTML$\to$TSV), Pseudo (Pseudo$\to$Code), Travel (Travel Planning) and Countd. (Countdown). Note that in these results, we average across all available generation settings in these results.}

\label{tab:sft_mixtures}
\begin{tabular}{lccccccc}
\toprule
\textbf{Model} (SFT Mixture) & HTML & Pseudo & Travel & Countd. & RAG & Rerank & Recall \\
\midrule
\bf Llama-3.1-8B-Base & 14.4 & 1.5 & 12.5 & 0.3 & 54.5 & 9.3 & 76.6 \\
$\;$ (UltraChat) & 24.4 & 4.0 & 10.0 & 21.7 & 57.5 & 15.1 & 88.2 \\
$\;$ (Tulu-v3) & 33.4 & 41.0 & 15.0 & 27.7 & 56.2 & 15.2 & 94.3 \\
$\;$ (Both, 1:1 ratio) & 32.6 & 41.5 & 16.0 & 30.3 & 57.4 & 17.3 & 95.6 \\
\addlinespace
\bf Llama-3.1-8B-Instruct & 33.0 & 47.5 & 27.5 & 7.7 & 59.5 & 14.0 & 95.2 \\
\midrule
\bf ProLong-8B-512K-Base & 31.0 & 44.0 & 7.5 & 35.7 & 59.6 & 17.4 & 97.8 \\
$\;$ (Ultrachat) & 34.6 & 12.5 & 6.0 & 8.7 & 64.5 & 21.2 & 99.0 \\
$\;$ (Tulu-v3) & 40.5 & 35.5 & 20.5 & 12.7 & 61.1 & 20.0 & 97.4 \\
$\;$ (Both, 1:1 ratio) & 42.8 & 33.0 & 15.5 & 27.7 & 63.1 & 20.8 & 99.4 \\
\bottomrule
\end{tabular}
\end{table}

\section{SFT training}
\label{app:sft}

We explore the interaction between PruLong and SFT training. For these experiments, we use the ProLong-8B-Base model~\citet{gao2025how}, for which both the long-context pre-training distribution (which is shared by our PruLong training) and the SFT data mixture is known.
\citet{gao2025how} find that short-context SFT data produces good long-context abilities after sufficient long-context pre-training. By default, the SFT dataset is UltraChat-200K~\citet{ding-etal-2023-enhancing}.
However, in exploratory experiments, we found that a mix of both UltraChat-200K and the Tulu-3-SFT mixture~\citep{lambert2025tulu3} produced slightly better downstream results when applied to ProLong-8B-Base, and for Llama-3.1-8B-Base, produced an instruction-tuned model almost on par with Llama-3.1-8B-Instruct, see Table~\ref{tab:sft_mixtures}
However, as we discovered the sensitivity to pre-filling chunk size when applying PruLong before SFT, our focus shifted to performing extended experiments on the Llama-3.1-8B-Instruct model, which no longer required us to find a stronger SFT setting.

The hyperparameters for the SFT training stage are provided in Table~\ref{tab:hyperparams1}. 

\section{Additional Results}
\label{app:results}
\begin{figure}[h]
    \centering
    \includegraphics[width=1.0\textwidth]{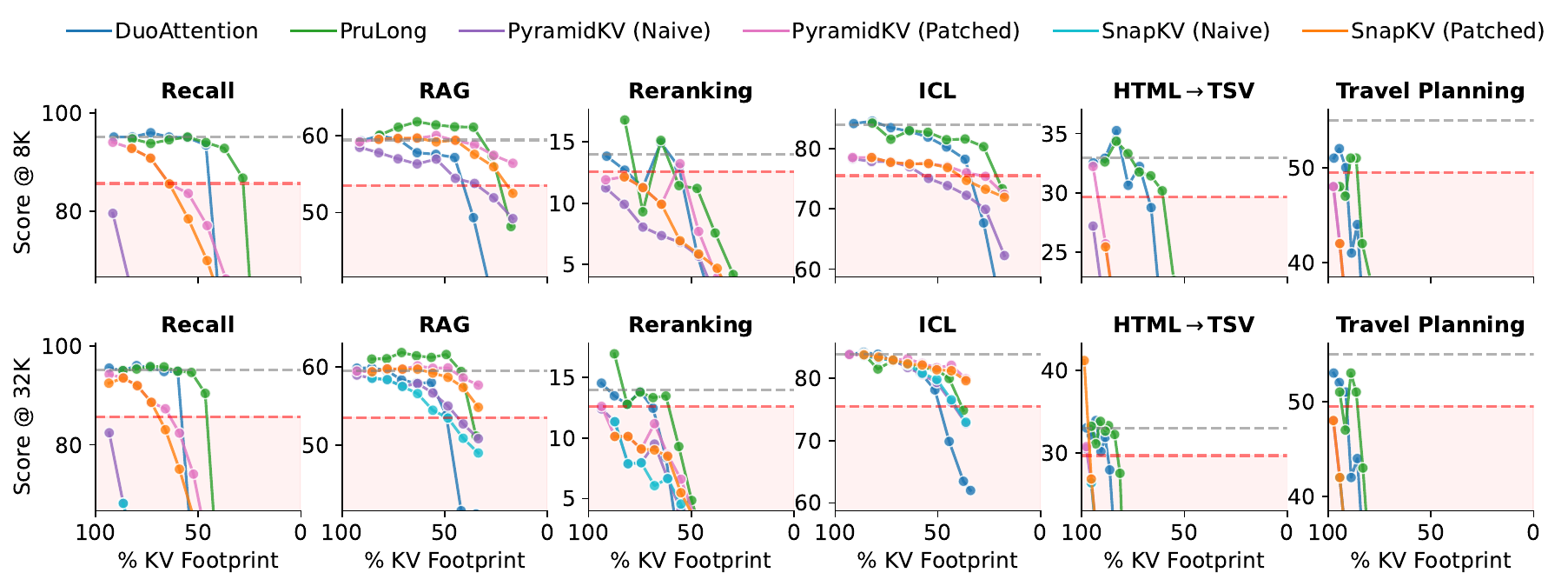}
    \caption{Performance vs. KV Footprint for the baselines and \ours{}. The gray dashed line denotes the original model's performance, and the red one represents $90\%$ of model performance. We show results at both 8K and 32K prefilling chunk sizes.}
    \label{fig:full_kv_footprint}
\end{figure}

In Section~\ref{sec:experiments}, we plotted the score of the Llama-3.1-8B-Instruct model on various task groups from HELMET and LongProc, at a 32K pre-filling chunk size.
In this appendix, we expand those results on two axes: we include results at a pre-filling chunk size of 8K tokens, and include another method: SnapKV.
Once again, we include a naive version of SnapKV and a patched version as per Section~\ref{sec:chunked_eviction}.
The expanded plots are displayed in Figure~\ref{fig:full_kv_footprint}.

In the 32K setting, we observe that PyramidKV almost always outperforms SnapKV.
Once again, the patched version of SnapKV achieves a higher score than the naive version at the same footprint.
Similar to the results of Appendix~\ref{app:kv_metrics}, we notice that PyramidKV and SnapKV suffer more when the pre-filling chunk size is shrunk to 8K tokens.
The effect is particularly true on Recall and ICL.

%% file: tables/tasks.tex
\begin{table}[t!]
    \caption{
        Overview of evaluation datasets. We use the tasks from HELMET~\citep{yen2025helmet} and LongProc~\citep{ye2025longprocbenchmarkinglongcontextlanguage}, but focus on task categories and generation settings where 8B parameter models attain non-trivial performance.
        ``\# Input'' and ``\# Output'' refer to the average number of input and output tokens respectively.
    }
    \centering
    \resizebox{0.98\linewidth}{!}{
        \begin{tabular}{lllp{0.45\linewidth}cc}
            \toprule
            \textbf{Category} & \textbf{Dataset} & \textbf{Metrics} & \textbf{Description} & \textbf{\# Input} & \textbf{\# Output} \\
            \midrule
            \multicolumn{6}{c}{\textbf{HELMET}} \\
            \midrule
           \multirow{6}{7em}{\textbf{Recall}} & JSON KV & SubEM & Retrieve a key in JSON  \citep{Liu2023LostIT} & 91K & 100 \\
            & RULER MK Needle & SubEM & Retrieve the needle (a number) within noisy needles \citep{hsieh2024ruler}& 95K & 6 \\
            & RULER MK UUID & SubEM& Retrieve the needle (a UUID) within noisy needles & 93K & 40  \\
            & RULER MV & SubEM & Retrieve multiple values for one needle (key) & 117K & 50 \\
            \midrule
            \multirow{4}{7em}{\textbf{RAG}}& Natural Questions & SubEM & Factoid QA \citep{kwiatkowski2019natural}   & 121K & 20 \\
            & TriviaQA & SubEM & Trivia QA \citep{joshi2017triviaqa} & 121K & 20 \\
            & PopQA & SubEM & Long-tail entity QA \citep{mallen-etal-2023-trust} & 113K & 20 \\
            & HotpotQA & SubEM & Multi-hop QA \citep{yang-etal-2018-hotpotqa} & 121K & 20 \\
            \midrule
            \textbf{Re-ranking} & MS MARCO & NDCG@10 & Rerank passage for a query~\citep{bajaj2016ms} & 85K & 200 \\
            \midrule
            \multirow{9}{6em}{\textbf{Many-shot in-context learning (ICL)}} & TREC Coarse  & Accuracy & Question  classification, 6 labels \citep{li-roth-2002-learning} & 106K & 20 \\
            & TREC Fine & Accuracy & Question  classification, 50 labels  & 104K & 20 \\
            & NLU & Accuracy & Task intent classification, 68 labels \citep{liu2021benchmarking} & 107K & 20 \\
            & BANKING77 & Accuracy & Banking intent classification,  77 labels \citep{casanueva-etal-2020-efficient} & 108K & 20 \\
            & CLINC150 & Accuracy & Intent classification, 151 labels ~\citep{larson-etal-2019-evaluation}& 106K & 20 \\
            \midrule
            \multirow{2}{6em}{\textbf{Long-document QA}} & NarrativeQA & Model-based & Book/movie script QA \citep{kocisky2018narrativeqa} & 112K & 25 \\
            & $\infty$~QA & ROUGE F1 & Novel QA with entity replacement \citep{zhang-etal-2024-bench} & 109K & 7 \\
            \midrule
            \multirow{2}{6em}{\textbf{Summarization}} & $\infty$~Sum & Model-based & Novel summarization with entity replacement & 108K & 810 \\
            & Multi-LexSum & Model-based & Summarizing multiple legal documents \citep{multilexsum} & 105K & 400 \\
            \midrule
            \multicolumn{6}{c}{\textbf{LongProc}} \\
            \midrule
            \textbf{HTML$\to$TSV} & -& F1 (row) & Extract website info into TSV & 12K & 1K \\
            & & & Averaged over three input/ouput lengths & 24K & 3K  \\
            & & & & 38K & 10K  \\
            \textbf{Travel Planning} &- & Accuracy & Generate multi-city itineraries under constraints~\citep{zheng2024natural}. We only 
 use the 6K→3K setting. & 6K & 3K \\
            \bottomrule
        \end{tabular}}
    \label{tab:dataset_overview}
\end{table}

%% file: appendices/real_metrics.tex
\section{Hardware metrics}
\label{ap:real_metrics}

\paragraph{Measurements} In this Appendix, we augment our comparison of the various methods on our idealized metrics (KV footprint and peak KV) with a comparison on the basis of ``real'' metrics---peak GPU memory utilization and throughput.
We measure the former in GiBs of memory, and the latter in terms of the number of requests per second.
The results for the various tasks are reported in Tables~\ref{tab:recall} through~\ref{tab:travel-planning}.
We also ablate our two modifications to PyramidKV and SnapKV: (1) the mean-pooling of attention within KV groups to avoid replicating KVs (C), and (2) patching in chunked eviction (P).
DuoAttention and PruLong were run with a head sparsity of $70\%$ (corresponding to a KV footprint slightly above $30\%$), and PyramidKV and SnapKV were allowed a total cache size of $30\%$ of the context window in these experiments---this leads to a KV footprint around $35\%$ in most cases.

We make the following observations:
\begin{enumerate}
    \item DuoAttention and PruLong usually achieve the highest throughput and the lowest peak memory. 
    \item Mean-pooling (C) usually reduces peak memory utilization by around $25\%$, but reduces throughput slightly for PyramidKV and SnapKV. It does not usually affect performance.
    \item Patching (P) leads to a higher score in most settings, and does not substantially affect the throughput or peak memory. The combination of P + C is usually the best-performing variant in both the PyramidKV and SnapKV groups.
    \item The ranking of methods generally tracks with those in Section~\ref{sec:experiments}, demonstrating that our idealized metrics lead to reliable takeaways. On the other hand, the precise values of the real metrics are noisy and show some variation across different runs.
\end{enumerate}

\paragraph{Why not just use hardware metrics then?} 
We saw above that our idealized metrics roughly track with hardware metrics.
One could then attempt to compare methods based on these hardware metrics alone.
Unfortunately, these hardware metrics are often confounded by differences between the implementations of different algorithms.
For instance, one implementation might utilize an optimized CUDA kernel, whereas another might not.
Another factor is that PyTorch often performs delayed garbage collection, which can confuse measurements of peak utilization.
Fixes like repeatedly clearing the CUDA cache might affect throughput instead.
There's also the issue of there not being a single ascribable value of peak memory and throughput---for instance, reducing peak memory by $5\%$ might suddenly allow a method to fit batch sizes twice as large, which may then translate to a higher throughput.
In essence, our idealized metrics represent the best achievable values of these real metrics---those that may be achieved with the optimal kernels and implementation, in theory.
Focusing on this ideal allows us to ignore all the implementational differences mentioned above.

\begin{table}[h]
\centering
\caption{Real metrics for Recall. C denotes compression by mean-pooling the attention from different queries in the same KV group. P denotes patched eviction.}
\label{tab:recall}
\begin{tabular}{l c c c}
\toprule
\textbf{Method} & \textbf{Throughput ($\times10^{-2}$\,s$^{-1}$)} & \textbf{Peak Memory (GiB)} & \textbf{Score} \\
\midrule
DuoAttention   & 10.0 & 26.6 & 59.75 \\
PruLong        & 10.8 & 26.3 & 92.50 \\
\midrule
PyramidKV      &  7.8 & 43.8 & 28.63 \\
\quad + C      &  7.6 & 33.7 & 34.81 \\
\quad + P      &  8.1 & 43.8 & 74.06 \\
\quad + P + C  &  8.0 & 33.7 & 88.00 \\
\midrule
SnapKV         &  8.0 & 43.7 & 24.94 \\
\quad + C      &  8.6 & 33.7 & 32.50 \\
\quad + P      &  8.2 & 43.8 & 65.38 \\
\quad + P + C  &  8.2 & 33.7 & 79.25 \\
\bottomrule
\end{tabular}
\end{table}

\begin{table}[h]
\centering
\caption{Real metrics for RAG. C denotes compression by mean-pooling the attention from different queries in the same KV group. P denotes patched eviction.}
\label{tab:rag}
\begin{tabular}{l c c c}
\toprule
\textbf{Method} & \textbf{Throughput ($\times10^{-2}$\,s$^{-1}$)} & \textbf{Peak Memory (GiB)} & \textbf{Score} \\
\midrule
{}             & None & None & 59.46 \\
\midrule
DuoAttention   &  9.6 & 28.8 & 53.83 \\
PruLong        &  9.4 & 28.6 & 61.00 \\
\midrule
PyramidKV      &  8.1 & 46.4 & 55.00 \\
\quad + C      &  8.1 & 34.5 & 54.67 \\
\quad + P      &  8.2 & 46.4 & 59.88 \\
\quad + P + C  &  8.0 & 34.6 & 60.46 \\
\midrule
SnapKV         &  8.1 & 46.4 & 53.46 \\
\quad + C      &  7.6 & 34.5 & 53.25 \\
\quad + P      &  8.1 & 46.4 & 58.67 \\
\quad + P + C  &  7.9 & 34.6 & 58.79 \\
\bottomrule
\end{tabular}
\end{table}

\begin{table}[h]
\centering
\caption{Real metrics for Reranking. C denotes compression by mean-pooling the attention from different queries in the same KV group. P denotes patched eviction.}
\label{tab:reranking}
\begin{tabular}{l c c c}
\toprule
\textbf{Method} & \textbf{Throughput ($\times10^{-2}$\,s$^{-1}$)} & \textbf{Peak Memory (GiB)} & \textbf{Score} \\
\midrule
DuoAttention   &  5.8 & 24.5 &  1.20 \\
PruLong        &  5.8 & 24.3 &  8.28 \\
\midrule
PyramidKV      &  4.8 & 40.5 &  4.06 \\
\quad + C      &  4.2 & 33.3 &  2.99 \\
\quad + P      &  4.4 & 40.6 &  6.60 \\
\quad + P + C  &  4.1 & 33.4 &  5.79 \\
\midrule
SnapKV         &  4.5 & 40.5 &  4.55 \\
\quad + C      &  4.5 & 33.3 &  3.78 \\
\quad + P      &  4.3 & 40.6 &  5.48 \\
\quad + P + C  &  4.2 & 33.4 &  6.56 \\
\bottomrule
\end{tabular}
\end{table}

\begin{table}[h]
\centering
\caption{Real metrics for ICL. C denotes compression by mean-pooling the attention from different queries in the same KV group. P denotes patched eviction.}
\label{tab:icl}
\begin{tabular}{l c c c}
\toprule
\textbf{Method} & \textbf{Throughput ($\times10^{-2}$\,s$^{-1}$)} & \textbf{Peak Memory (GiB)} & \textbf{Score} \\
\midrule
DuoAttention   & 11.0 & 27.4 & 78.16 \\
PruLong        & 11.0 & 27.1 & 81.36 \\
\midrule
PyramidKV      &  9.6 & 45.3 & 79.36 \\
\quad + C      &  9.3 & 34.5 & 79.24 \\
\quad + P      &  9.4 & 45.4 & 81.72 \\
\quad + P + C  &  8.9 & 34.6 & 82.64 \\
\midrule
SnapKV         &  9.5 & 45.3 & 79.84 \\
\quad + C      &  9.1 & 34.5 & 78.80 \\
\quad + P      &  9.2 & 45.4 & 81.40 \\
\quad + P + C  &  8.7 & 34.6 & 81.64 \\
\bottomrule
\end{tabular}
\end{table}

\begin{table}[h]
\centering
\caption{Real metrics for HTML $\to$ TSV. C denotes compression by mean-pooling the attention from different queries in the same KV group. P denotes patched eviction.}
\label{tab:html-to-tsv}
\begin{tabular}{l c c c}
\toprule
\textbf{Method} & \textbf{Throughput ($\times10^{-2}$\,s$^{-1}$)} & \textbf{Peak Memory (GiB)} & \textbf{Score} \\
\midrule
DuoAttention   &  1.8 & 17.0 & 18.24 \\
PruLong        &  1.9 & 16.9 & 30.47 \\
\midrule
PyramidKV      &  1.1 & 22.0 &  1.43 \\
\quad + C      &  1.8 & 16.9 &  2.33 \\
\quad + P      &  1.1 & 22.0 &  1.05 \\
\quad + P + C  &  1.8 & 16.9 &  2.14 \\
\midrule
SnapKV         &  1.2 & 22.0 &  1.12 \\
\quad + C      &  1.3 & 19.0 &  1.50 \\
\quad + P      &  1.1 & 22.0 &  1.06 \\
\quad + P + C  &  1.3 & 19.0 &  1.62 \\
\bottomrule
\end{tabular}
\end{table}

\begin{table}[h]
\centering
\caption{Real metrics for Travel Planning. C denotes compression by mean-pooling the attention from different queries in the same KV group. P denotes patched eviction.}
\label{tab:travel-planning}
\begin{tabular}{l c c c}
\toprule
\textbf{Method} & \textbf{Throughput ($\times10^{-2}$\,s$^{-1}$)} & \textbf{Peak Memory (GiB)} & \textbf{Score} \\
\midrule
DuoAttention   &  1.2 & 16.1 & 22.00 \\
PruLong        &  1.5 & 16.1 & 37.00 \\
\midrule
PyramidKV      &  0.8 & 18.9 &  3.00 \\
\quad + C      &  1.1 & 17.1 &  4.00 \\
\quad + P      &  0.8 & 18.9 &  3.00 \\
\quad + P + C  &  1.1 & 17.1 &  4.00 \\
\midrule
SnapKV         &  1.0 & 18.9 &  5.00 \\
\quad + C      &  1.4 & 17.1 &  5.00 \\
\quad + P      &  1.1 & 18.9 &  5.00 \\
\quad + P + C  &  1.4 & 17.1 &  5.00 \\
\bottomrule
\end{tabular}
\end{table}